\documentclass[letterpaper]{article} % DO NOT CHANGE THIS
\usepackage{url}
\usepackage{aaai23}  % DO NOT CHANGE THIS
\usepackage{times}  % DO NOT CHANGE THIS
\usepackage{helvet}  % DO NOT CHANGE THIS
\usepackage{courier}  % DO NOT CHANGE THIS
\usepackage{graphicx} % DO NOT CHANGE THIS
\usepackage{natbib}  % DO NOT CHANGE THIS AND DO NOT ADD ANY OPTIONS TO IT
\usepackage{caption} % DO NOT CHANGE THIS AND DO NOT ADD ANY OPTIONS TO IT
\usepackage{algorithm}
\usepackage{algorithmic}
\usepackage{booktabs}
\usepackage[T1]{fontenc}
\usepackage[table,RGB]{xcolor}
\usepackage{amsmath,amsfonts,amssymb,mathtools}
\usepackage{amsthm} % custom for proof
\usepackage{bm}
\newtheorem{proposition}{Proposition}

\usepackage{cases}
\usepackage{comment}

\usepackage{animate}

\newcommand{\beas}{\begin{eqnarray*}}
\newcommand{\eeas}{\end{eqnarray*}}
\newcommand{\bea}{\begin{eqnarray}}
\newcommand{\eea}{\end{eqnarray}}
\newcommand{\bes}{\begin{equation*}}
\newcommand{\ees}{\end{equation*}}
\newcommand{\be}{\begin{equation}}
\newcommand{\ee}{\end{equation}}

\makeatletter
\usepackage{xspace}
\def\@onedot{\ifx\@let@token.\else.\null\fi\xspace}
\DeclareRobustCommand\onedot{\futurelet\@let@token\@onedot}

\newcommand{\figref}[1]{Fig\onedot~\ref{#1}}

\newcommand{\tabref}[1]{Table\onedot~\ref{#1}}

\usepackage[footnote,printonlyused]{acronym}
\usepackage{scrextend}
\deffootnote[1.0em]{3.2em}{0em}
{\textsuperscript{\thefootnotemark}\,\enskip}

\usepackage{newfloat}
\usepackage{listings}
\DeclareCaptionStyle{ruled}{labelfont=normalfont,labelsep=colon,strut=off} % DO NOT CHANGE THIS
\floatstyle{ruled}
\newfloat{listing}{tb}{lst}{}
\floatname{listing}{Listing}
\title{PowRL: A Reinforcement Learning Framework for Robust Management of Power Networks}
\author{
    Anandsingh Chauhan\textsuperscript{\rm 1},
    Mayank Baranwal\textsuperscript{\rm 1,2},
    Ansuma Basumatary\textsuperscript{\rm 3}
}
\affiliations{
    \textsuperscript{\rm 1}Tata Consultancy Services Research, Mumbai\\
    \textsuperscript{\rm 2}Indian Institute of Technology, Bombay\\
    \textsuperscript{3}SalesKen, Bengaluru\\
    anandsingh.chauhan@tcs.com,
    baranwal.mayank@tcs.com, ansuma.bty25@gmail.com
}

\begin{document}

\maketitle

\begin{abstract}
Power grids, across the world, play an important societal and economical role by providing uninterrupted, reliable and transient-free power to several industries, businesses and household consumers. With the advent of renewable power resources and EVs resulting into uncertain generation and highly dynamic load demands, it has become ever so important to ensure robust operation of power networks through suitable management of transient stability issues and localize the events of blackouts. In the light of ever increasing stress on the modern grid infrastructure and the grid operators, this paper presents a reinforcement learning (RL) framework, PowRL, to mitigate the effects of unexpected network events, as well as reliably maintain electricity everywhere on the network at all times. The PowRL leverages a novel heuristic for overload management, along with the RL-guided decision making on optimal topology selection to ensure that the grid is operated safely and reliably (with no overloads). PowRL is benchmarked on a variety of competition datasets hosted by the L2RPN (Learning to Run a Power Network). Even with its reduced action space, PowRL tops the leaderboard in the L2RPN NeurIPS 2020 challenge (Robustness track) at an aggregate level, while also being the top performing agent in the L2RPN WCCI 2020 challenge. Moreover, detailed analysis depicts state-of-the-art performances by the PowRL agent in some of the test scenarios.
\end{abstract}

\section{Introduction}
The infrastructure that defines the modern electricity grid is largely based on centralized power generation units, such as the fossil-fuel-fired power plants and nuclear power plants~\cite{wulf2000great}. Through out its evolution, the primary focus of grid development has largely been around safety, reliability, and resiliency to uncertain events. The ever increasing penetration of centralized renewable generation units, such as the hydroelectric dams, as well as the distributed energy resources (DERs), such as the solar or wind farms, has resulted in significant variability in power generation. Moreover, the customers' evolving energy usage patterns through charging of electric vehicles (EVs), batteries, and accessibility to modern electrical appliances have furthered the uncertainty and variability in the electric grid, not to mention the occasional adversarial cyberattacks in order to disrupt reliable power delivery by causing blackouts.

Modernization of the electricity grid through innovative hardware and software solutions involving peer-to-peer networks of power electric converters for enabling coordinated response to adjust power generation and consumption can partially alleviate some of the challenges~\cite{baranwal2018distributed,arwa2020reinforcement}. However, overhauling the entire grid infrastructure, unfortunately, operates at its own time-scale. This work addresses the challenge of leveraging AI-driven solutions for extracting the maximum possible resiliency out of the current grid infrastructure in order to ensure smooth power delivery and provide ancillary services to the grid operator. The challenge lies not only in dealing with the uncertainty of power demand and generation, or the uncertain events, such as electrical faults or adversarial attacks on the grid, but also with the huge (combinatorially large) action space even in a moderately-sized grid. From the perspective of the grid operator, devising a real-time strategy for the robust management of power networks is beyond human cognition. In most such scenarios, the grid operator relies on his/her own experience or at best, some of the potential heuristics whose scope is limited to mitigating only a certain types of uncertainties.

\begin{figure*}
	\begin{center}
		\begin{tabular}{c}
			\includegraphics[width=1.9\columnwidth]{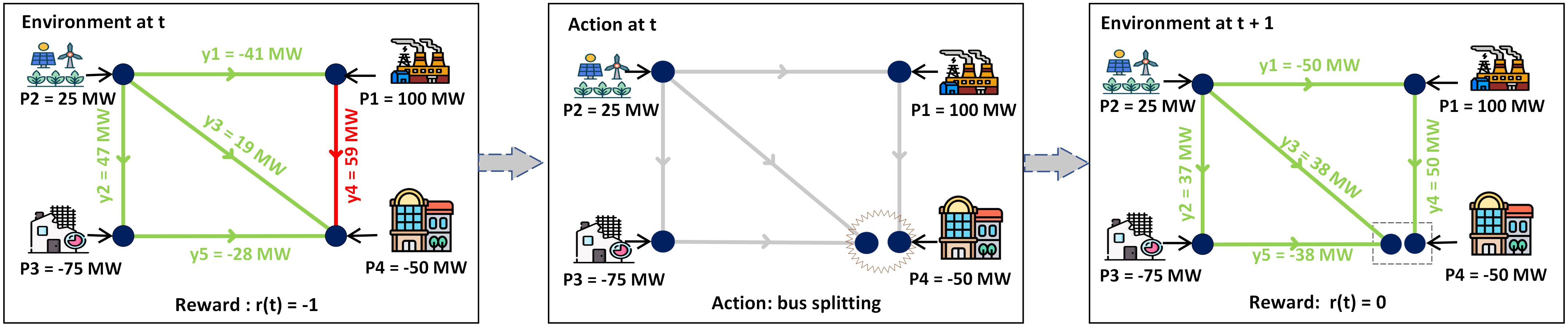}
		\end{tabular}
		\caption{Reliable operation through bus-splitting: instead of disconnecting the overloaded line, a simpler and effective solution is to alter the underlying network topology.}
		\label{fig:mot-example}
		\vspace{-.5em}
	\end{center}
	\vspace{-1.5em}
\end{figure*}

%\noindent {\bf Motivating example} \figref{fig:mot-example} represents a typical scenario in a power network, where one of the transmission lines (shown in red) is overloaded beyond its thermal capacity. If left unaccounted, the overflow may result in permanent damage of the transmission line resulting in significant delays in getting the network back to its nominal state. One can potentially mitigate the overflow by modifying the power generation (redispatch) at a neighboring bus, or even disconnect the power line manually to avoid any permanent damage. While both these actions are valid control actions, effects of generator redispatch and line disconnection followed by reconnection after cooldown, evolve at a much slower time-scale. Interestingly, an optimal and real-time control strategy is to identify substations where a bus-split action (as shown in the rightmost grid) results in immediate mitigation of line overflow through appropriate changes in the underlying network topology. Finding the most-suited topology control action among the list of all possible topologies is not straightforward, especially if the power network is equipped with many substations with each substation consisting of multiple control elements (lines, generators and loads).
\noindent {\bf Motivating example} \figref{fig:mot-example} represents a typical scenario in a power network, where one of the transmission lines (shown in red) is overloaded beyond its thermal capacity. If left unaccounted, the overflow may result in permanent damage of the transmission line resulting in significant delays in getting the network back to its nominal state. Interestingly, an optimal and real-time control strategy is to identify substations where a bus-split action (as shown in the rightmost grid) results in immediate mitigation of line overflow through appropriate changes in the underlying network topology. Finding the most-suited topology control action among the list of all possible topologies is not straightforward, especially if the power network is equipped with many substations with each substation consisting of multiple control elements.

% \noindent {\bf Why RL?} The penetration of DERs, EVs, and complex connectivities between different substations is increasingly rendering the traditional control strategies used by the electrical engineers inadequate. On the other hand, recent advancements in machine learning (ML) may offer solutions to such complex problems which are otherwise beyond human comprehension. A key feature of most power systems is that the operating point, characterized by the net power generation and load demand, is highly dynamic. Consequently, managing a power network reliably entails real-time sequential decision making, for which reinforcement learning (RL) is most suited. In fact, the underlying dynamics of a power network is specified using the fundamental laws of electricity, which are typically nonlinear. Finding an optimal sequence of actions using an integer program (or related optimization framework) that lead to a resilient grid performance are suitably intractable. Given the potential solutions that an RL-based framework offers, the ``Learning to run a power network" (L2RPN) challenge was recently conceptualized to model the sequential decision-making environments of real-time power network operations. The challenge introduced a realistically-sized network environment (based on the IEEE-118 network), which is a reduced-order approximation of the American Electric Power system (in the U.S. Midwest). L2RPN is aimed at testing the feasibility of an RL-based approach towards developing a smart operation recommender system for grid operators.\vspace{-.5em}

\noindent {\bf Why RL?} The penetration of DERs, EVs, and complex connectivities between different substations is increasingly rendering the traditional control strategies used by the electrical engineers inadequate. On the other hand, recent advancements in machine learning (ML) may offer solutions to such complex problems which are otherwise beyond human cognition. A key feature of most power systems is that the operating point, characterized by the net power generation and load demand, is highly dynamic. Consequently, managing a power network reliably entails real-time sequential decision making, for which reinforcement learning (RL) is most suited. Given the potential solutions that an RL-based framework offers, the ``Learning to run a power network" (L2RPN) challenge was recently conceptualized to model the sequential decision-making environments of real-time power network operations. The challenge introduced a realistically-sized network environment (based on the IEEE-118 network), which is a reduced-order approximation of the American Electric Power system (in the U.S. Midwest). L2RPN is aimed at testing the feasibility of an RL-based approach towards developing a smart operation recommender system for grid operators.\vspace{-.5em}

\subsection{Statement of Contributions}
In this work, we introduce an RL-based recommender system, titled \emph{PowRL}, for robust management of power networks subjected to adversarial attacks on transmission lines (uncertain events), as well as temporal variations in demand and supply. PowRL is built upon the \emph{Grid2Op}~\cite{grid2op} framework that facilitates users to plug in the underlying simulator of a model of their choice. Below we summarize the primary contributions of our work.
\begin{enumerate}
    \item {\bf Grid management with reduced action-space} The total number of feasible topology actions in the reduced IEEE-118 system are in the order of 70k. Working with such a large action space is beyond the scope of any practical RL-agent. Following the work in~\cite{zhou2021action,yoon2021winning}, we identify important substations and a significantly reduced action set (comprising of 240 unique actions) through extensive simulations in order to facilitate faster learning by agents.\vspace{-.25em}
    \item {\bf Combining heuristics with deep-learning} Traditional control of electrical grids is often based on a supervisory rule-based strategy. We augment a novel rule-based scheme with the RL agent during both training and inference phases in order to develop a coordinated control response to uncertain events (see \figref{fig:flowchart}).\vspace{-.25em}
    \item {\bf SOTA performance on benchmark datasets} PowRL is benchmarked on several L2RPN competition datasets, and is shown to significantly outperform the classical heuristics, as well as other RL agents.\vspace{-.25em}
    \item {\bf Post-hoc analysis and insights} We also analyze the sequences of control actions adopted by the PowRL agent across various scenarios, and observe that unlike other RL agents, the action sequences of PowRL are more diversified suggesting improved learning behavior.
\end{enumerate}

\section{Preliminaries}
A power network comprises of several \emph{substations} connected with each other through transmission \emph{lines}. Each substation is equipped with electrical elements, such as \emph{generators} and/or \emph{loads}. For instance, the network in \figref{fig:mot-example} consists of 4 substations (nodes), 5 transmission lines (edges), 2 generators ($P1$ and $P2$), and 2 loads ($P3$ and $P4$). The generators produce power, only to be consumed by the network loads and through transmission losses. The power lines are subjected to maximum current (power) carrying capacities, and an excess flow through power lines for a sizable duration results in permanent damage and eventual disconnection from the network. A substation acts as a router in the network and determines where to transmit power. Additionally, a substation is also equipped with multiple conductors, known as \emph{buses}. The last subfigure in \figref{fig:mot-example} represents a scenario where the network at the fourth substation is routed through 2 separate buses.\vspace{-.5em}
\begin{figure}
    \centering
    \includegraphics[width=.99\columnwidth]{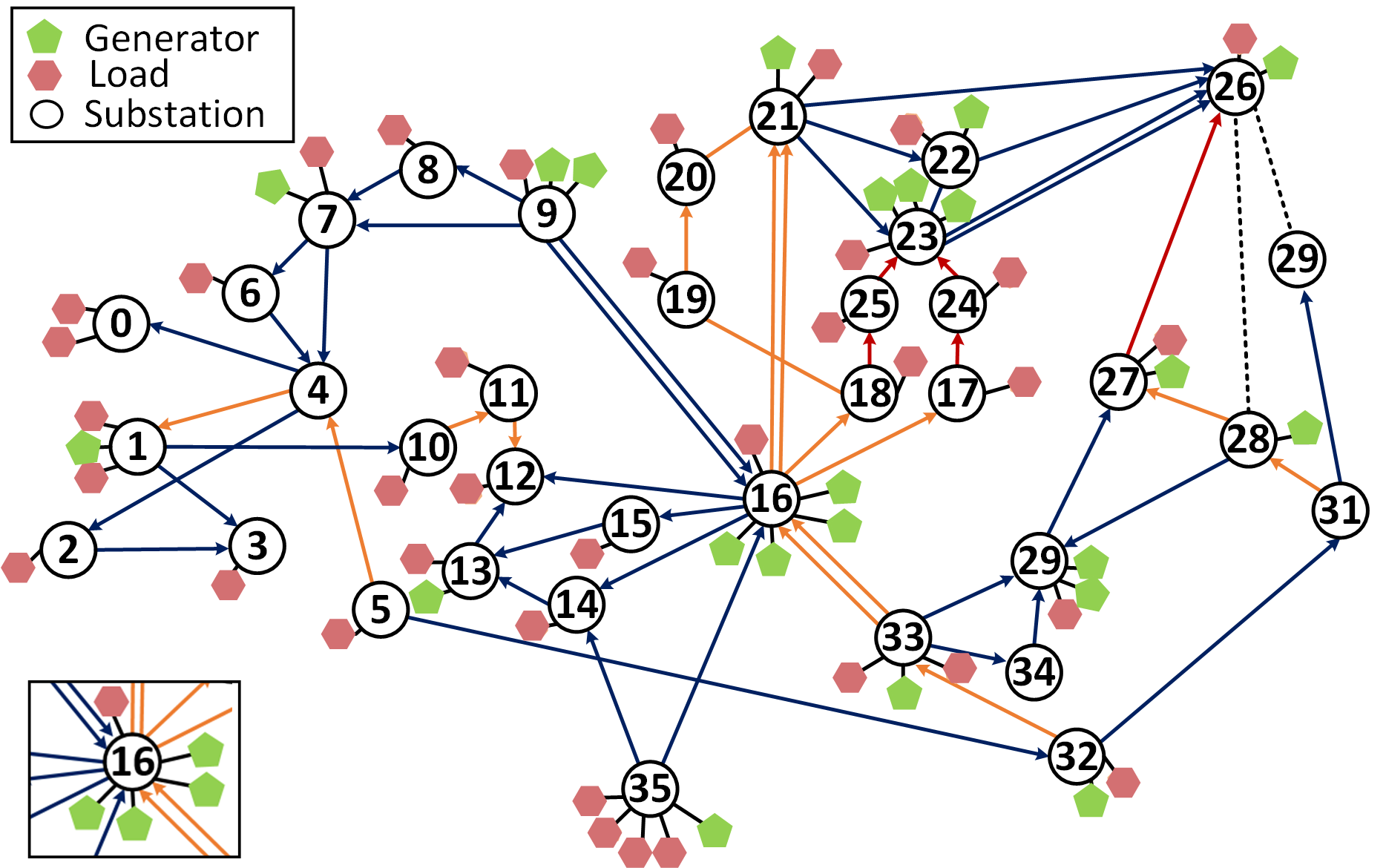}
    \caption{Synthetic IEEE-118 network consisting of 36 substations, 59 lines, 22 generators.}
	\label{fig:PowGrid}
	\vspace{-1.5em}
\end{figure}

\subsection{The Grid2Op Environment}
The L2RPN challenge is built upon an open-source simulator, \emph{Grid2Op}, for power grid operation. Grid2Op offers flexibility to work with realistic scenarios, execute contingency events, and test control algorithms subjected to several physical and operational constraints. The challenge scenario consists of industry standard synthetic IEEE-118 network with 36 substations, 59 lines and 22 generators (see \figref{fig:PowGrid}). The remainder of the IEEE-118 network is represented as a time-varying load. Grid2Op considers 2 buses per substation, also known as the double busbar system. Grid2Op environment is also equipped with realistic production and consumption scenarios, as well as adversarial attacks on the network manifested as forced line failures. The agents can take remedial actions subjected to following constraints:\\
\noindent (a) Deterministic events, such as maintenance, and adversarial events, such as meteorological conditions, can disconnect lines for substantial duration\\
\noindent (b) Each power line has a maximum flow capacity, and gets disconnected automatically if overloaded for too long\\
\noindent (c) Disconnected lines cannot be immediately reconnected\\
\noindent (d) Agents cannot act too frequently on the same line, generator or substation in order to avoid asset degradation\\
The episode terminates when the total load demand is not met or during incidents of wide-area blackout\vspace{-.5em}

\subsection{States, Actions and Rewards}
The problem of controlling power networks can be cast as a Markov Decision Process (MDP). An MDP is denoted by the tuple $\langle\mathcal{S},\mathcal{A},\mathcal{P}_\mathcal{A},r\rangle$, where 
$\mathcal{S}$ and $\mathcal{A}$ represent the finite set of states and actions, respectively. For each $s,s'\in\mathcal{S}$, the probability of transition from $s\to s'$ under the effect of an action $a\in\mathcal{A}$, is denoted by $p_a(s,s')\in\mathcal{P}_\mathcal{A}$. Finally, the step-reward associated with each state-action pair $(s,a)$ is depicted by $r(s,a)$. Below, we summarize the set of all possible states, actions and reward:\\
\noindent\textbf{States} Agents can access the entire state of the power network at each time-step, including the demand forecast at the next step, load flow and status of power lines, voltages at each busbar, production at each generator, and various operational constraints\\
\noindent\textbf{Actions} Grid2Op is equipped with two kinds of actions: (a) combinatorial (or discrete), (b) continuous. Discrete actions are related to inexpensive topological actions, such as line disconnection/reconnection, or actions at the busbar. Additionally, the generators can be redispatched through predefined continuous actions using costly production changes. For the reduced IEEE-118 system, there are nearly 70k discrete actions and 40 continuous actions.\\
\noindent\textbf{Rewards} The L2RPN challenge is equipped with a competition specific reward, however, Grid2Op also facilitates inclusion custom reward functions. More details on the exact reward structure are included in~\cite{marot2021learning}.

\section{Challenges in Controlling Power Network}
\noindent\textbf{Combinatorially many actions} The primary challenge with executing remedial topological reconfiguration actions is that the number of discrete actions at a substation scales exponentially with the number of elements connected to the substation. For instance, the substation~16 in \figref{fig:PowGrid} comprises of 17 elements (12 lines, 4 generators, 1 load). The number of topologies at this substation for a double busbar system is 65,505. The Proposition below enumerates all valid topologies for a double busbar substation.\vspace{-.25em}
\begin{proposition}\label{prop:total}
    The number of valid topological reconfigurations for a double busbar substation comprising of $N_{\text{line}}$ lines, $N_{\text{g}}$ generators, and $N_{\text{load}}$ loads is $2^{N_{\text{tot}}-1} - 2^{N_{\text{g}}+N_{\text{load}}} + 1$, where $N_{\text{tot}}\coloneqq N_{\text{line}}+N_{\text{g}}+N_{\text{load}}$.
\end{proposition}
\noindent\emph{Proof}: See Technical Appendix for details.

\noindent\textbf{Uncertainty with renewables} Renewable energy sources, such as solar or wind, depend heavily on the weather conditions. Thus, while the current topology may be adept at handling the net load demand, a significant change in power generation due to change in weather conditions may force the operator to take immediate remedial action in order to avoid transmission loss failures or eventual blackouts.\\
\noindent\textbf{Adversarial attacks} The L2RPN challenge operates with a heuristically designed opponent to mimic the $N-1$ security criterion in power networks~\cite{zhao2018graph}; it acts in an environment parallel to the RL agent and affects the power grid through forced contingencies, by disconnecting some of the targeted tensed lines at random times. The lines remain disconnected for maintenance and the agent can only reconnect them once the maintenance is over.

\section{Related Works}
%The aim of this study is to provide effective contingency management in power systems, which may be weather related (such as, reduced generation due to less wind or sun), line failures due to meteorological conditions, sudden changes in load demand, or cyberattacks on parts of the network. Grid operator must immediately respond with \emph{remedial} action in order to avoid potential line overflow or wide-area blackouts.
Among the most standard practices for managing contingency events is the `$N-1$'-criterion~\cite{ren2008long}, which requires that the network should continue to operate in the safe state even if one of the elements  (productions, lines, transformers, etc.) is disconnected. The criterion is usually enforced through solving a complex optimization problem that encode several physical constraints, as well as the fundamental laws of electricity~\cite{wang2013risk,dehghanian2015flexible,alhazmi2019power}. Due to the limitations with the traditional approaches, there is a shift towards adopting AI guided approaches as potential solution methodologies for robust control of power networks. One of the foundational works along this direction~\cite{donnot2017introducing} introduces a deep-learning based approach for remedial actions comprising of line disconnection/reconnection in a fixed-topology network. As such, their approach does not explore combinatorial topological actions. In the subsequent works, the authors proposed an expert system for topological remedial actions~\cite{marot2018expert} for the IEEE-14 system, which was further improved using a simple RL-based framework~\cite{marot2020learning} through introduction of the Grid2Op simulator and the first power grid management competition in 2019.

The winner of the first challenge~\cite{lan2020ai}, used pre-training to generate a good optimal control policy, followed by guided exploration in the large action space to select a top-few legal control actions at each time step through extensive simulations. They designed a dueling deep Q-network (DDQN)~\cite{wang2016dueling} with prioritized replay buffer~\cite{schaul2016pri}, only to act in contingency situations. Authors in~\cite{yoon2021winning} adopted a graph neural network (GNN) based actor-critic method to find the goal topology of the network given the current state. Their solution, though, was ranked first in the L2RPN WCCI challenge, did not perform nearly as well when applied to more complex scenarios as part of the L2RPN NeurIPS challenge.

The last L2RPN NeurIPS 2020 challenge saw new RL-based solutions that were significant improvements over the previously reported approaches~\cite{marot2021learning}. The third-place winner, \textbf{lujixiang}~\cite{lujixiang}, used a DDQN approach for topological actions, very similar to~\cite{lan2020ai}, however, with a reduced action space of 885 actions. The authors also included a base agent to handle remedial actions involving line disconnection/reconnection and generator redispatch. The second-place winner, \textbf{binbinchen}~\cite{binbinchen}, used a novel `Teacher-Tutor-Junior-Senior-Student' framework for warm-start their proximal policy optimization (PPO)~\cite{schulman2017proximal} model comprising of 208 unique remedial actions identified through extensive simulations. The top performing entry, \textbf{rl\_agent}, from Baidu~\cite{zhou2021action} relied on their PARL framework and Baidu's computational resources to train a very deep policy using evolutionary black-box optimization~\cite{lee2020efficient}.% on 5000 cores in 30 minutes.

\section{Our Approach: PowRL}
\begin{figure*}
	\begin{center}
		\begin{tabular}{c}
			\includegraphics[width=1.95\columnwidth]{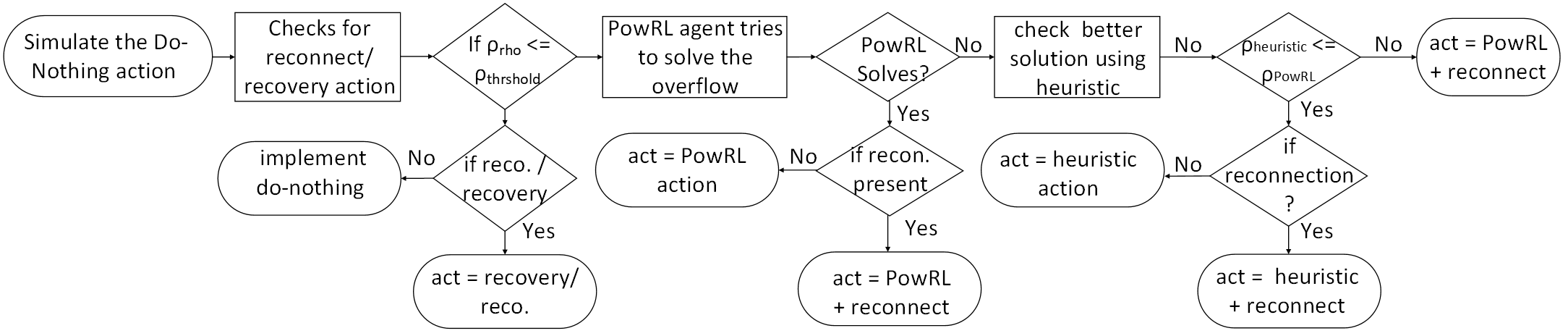}
		\end{tabular}
		\caption{Schematic of PowRL: PowRL combines RL-agent with a threshold based heuristic scheme}
		\label{fig:flowchart}
	\end{center}
	\vspace{-.5em}
\end{figure*}

\begin{table*}
    \centering
    \begin{tabular}{cccccc}
    	\toprule
    	Method & NeurIPS (Online) & NeurIPS (Offline) & WCCI & \# actions $(|\mathcal{A}|)$ & Run-time (s)\\
    	\midrule
    	\texttt{Do\_Nothing} & 0 & 0.34 & 50.09 & NA & 0.54\\
    	\texttt{expert\_heuristic} & 22.53 & 51.44 & 50.1 & NA & 9.76\\
    	\arrayrulecolor{black!30}\midrule
    	\texttt{lujixiang} & 45 & 53.96 & NA & 885 & 824.58\\
    	\texttt{binbinchen} & 52.42 & 3.96 & 51.1 & 208 + 1255 & 503.88\\
    	\texttt{rl\_agent} & 61.05 & {\bf 61.06} & {\bf 96.4} & 232 + 500 + redispatch & 358.82\\
    	\texttt{PowRL (Ours)} & {\bf 61.48} & 59.69 & {\bf 96.4} & {\bf 206 + 34} & {\bf 353.36}\\
    	\arrayrulecolor{black}\bottomrule
    \end{tabular}
    \caption{Performance on L2RPN Challenge Datasets}
    \label{tab:comp}
    \vspace{-1.5em}
\end{table*}

\subsection{Obtaining Reduced Action Space} \vspace{-.25em} Even with AI-guided solutions, working with very large action space (comprising of nearly 70k actions) is impractical. More importantly substations, such as the ones indexed by 9, 16, 23 and 26 (\figref{fig:PowGrid}) consist of relatively larger number of control elements, and understandably offer greater flexibility in terms of control of power networks. Through extensive simulations, we identify a much smaller of 240 actions that appeared to have most impact on network control, i.e., $|\mathcal{A}|=240$. The action set consists of only topological control. Costly generation production actions are excluded in our approach, as the inexpensive topological changes alone responded well to several contingency events.\vspace{-.25em}
\subsection{Heuristic-Guided Topological Actions} \vspace{-.25em} Other than building an RL agent to select a top-performing (electrically) valid action from the set $\mathcal{A}$, we also introduced some baseline heuristics to support the RL agent. These heuristics include: (a) disengage RL agent during `acceptable' grid operation, (b) allowing PowRL to reconnect a line back soon after its cooldown period ends, (c) any network reconfigurations are restored to original state as soon as the contingency ends, (d) reconnect the line back soon after the scheduled maintenance period is over, (e) do not disconnect lines that result in network bifurcation.\vspace{-.25em}
\subsection{Proximal Policy Optimization (PPO)-Guided Reinforcement Learning Framework} \vspace{-.25em} In order to learn to perform optimal sequential decision making, we train an on-policy PPO agent with a prioritized replay buffer defined over the set of network states. The state of the agent $S_t\in\mathcal{S}$ at time-step $t$ consists of: (a) the time-step information $t$ including the month, date, hour, minute and day of week information. (b) generator features $F_{\text{gen}}$ comprising of active $(P_{\text{gen}})$ and reactive $(Q_{\text{gen}})$ generation, and the voltage magnitude $V_{\text{gen}}$. (c) load features $F_{\text{load}}$ comprising of active $(P_{\text{load}})$ and reactive $(Q_{\text{load}})$ consumption, and the voltage magnitude $V_{\text{load}}$. (d) line features $F_{\text{line}}$ which include active $(P_{\text{or}},P_{\text{ex}})$ and reactive $(Q_{\text{or}},Q_{\text{ex}})$ power flows, as well as the bus voltages $(V_{\text{or}},V_{\text{ex}})$ and current flow $(a_{\text{or}},a_{\text{ex}})$, at both ends of the line. The other important line features are the power flow capacity $\rho$ a ratio to capture the amount of overflow in a line, timestep overflow $t_{\text{of}}$ to depict the duration of time steps since a line has overflowed, and line status $I_\text{l}$ to indicate if the line is connected or disconnected in the network. (e) Additionally, the state space also includes the topological vector status $I_{\text{topo\_vect}}$ to indicate on which bus each object of the power grid is connected in the substation. (f) The grid operator may have to wait for some time steps before a line or a substation can be acted upon, denoted by the line cooldown time step $t_\text{l}$ and substation cooldown time step $t_{\text{s}}$. (g) Power lines can go under some planned maintenance; the information of the time of next planned maintenance $t_{\text{nm}}$ and its duration $t_{\text{d}}$ is also included in the input features, and as an agent can’t operate on lines under maintenance. In summary, the state of the PPO agent is given by:
\[S_t \coloneqq [t,F_{\text{gen}},F_{\text{load}},F_{\text{line}},I_\text{l},\rho,I_{\text{topo\_vect}},t_{\text{of}},t_{\text{l}},t_{\text{s}},t_{\text{nm}},t_{\text{d}}].
\]
The step reward for training the PPO agent is designed to incur  additional penalty when the maximum overflow ratio $\rho_{\max}$ is beyond the safety threshold of 0.95, i.e.,
\[r = \left\{\begin{array}{cl}
     2 - \rho_{\max}, & \text{if} \ \rho_{\max} < 0.95, \\
     2 - 2\rho_{\max}, & \text{else}.
\end{array}\right.\]
Additionally, we incentivize the agent through an episodic reward of 500 upon surviving the entire episode. Premature termination of the episode due to any illegal action, or grid blackout is penalized through an episodic reward of -300.\vspace{-.25em}
\subsection{Optimal Action Selection} \vspace{-.25em}Controlling a power network using random exploratory actions may result in power grid collapse/blackout within a few steps. On the other hand, the \texttt{do-nothing} action that does not make any modifications to the existing grid topology or to the generator production, is no good either. In fact, the expected survival rate with the do-nothing action on the L2RPN NeurIPS 2020 challenge set is only $\sim24\%$. Recall that under nominal operating conditions, it is preferable to continue to operate with the starting topology. This makes it challenging to decide when to sample actions from the RL agent so that the deviation from the initial grid topological configuration is the least for better survival. We achieve this by combining heuristics with the action recommendation by the RL agent. \figref{fig:flowchart} explains the structured optimal action selection of the PowRL framework. Given the current state $S_t$ of the power system and overflow status of all power lines, the agent first simulates (not implements) the \texttt{do-nothing} action and obtains the simulated $\rho_{\text{do-nothing}}$. If $\rho_{\text{do-nothing}}$ is within the safety threshold, then the agent implements either the \texttt{do-nothing} action or the \texttt{recovery} action in case the current topology is different from the initial topology. In an event of overflow with $\rho_{\text{do-nothing}}$ exceeding the safety threshold, the PowRL framework implements the action recommended by the RL agent in conjunction with any possible line reconnection action. The performance of the PowRL agent depends on the choice of safety threshold $\rho_{\text{threshold}}$. In scenarios where the network is under severe attack from the adversaries, the PowRL framework needs to perform aggressive topological reconfiguration actions by frequently invoking the RL agent. This can be achieved by setting a smaller value of the safety threshold.

\section{Experiments}
\begin{figure*}
	\begin{center}
		\begin{tabular}{cc}
			\includegraphics[width=.82\columnwidth]{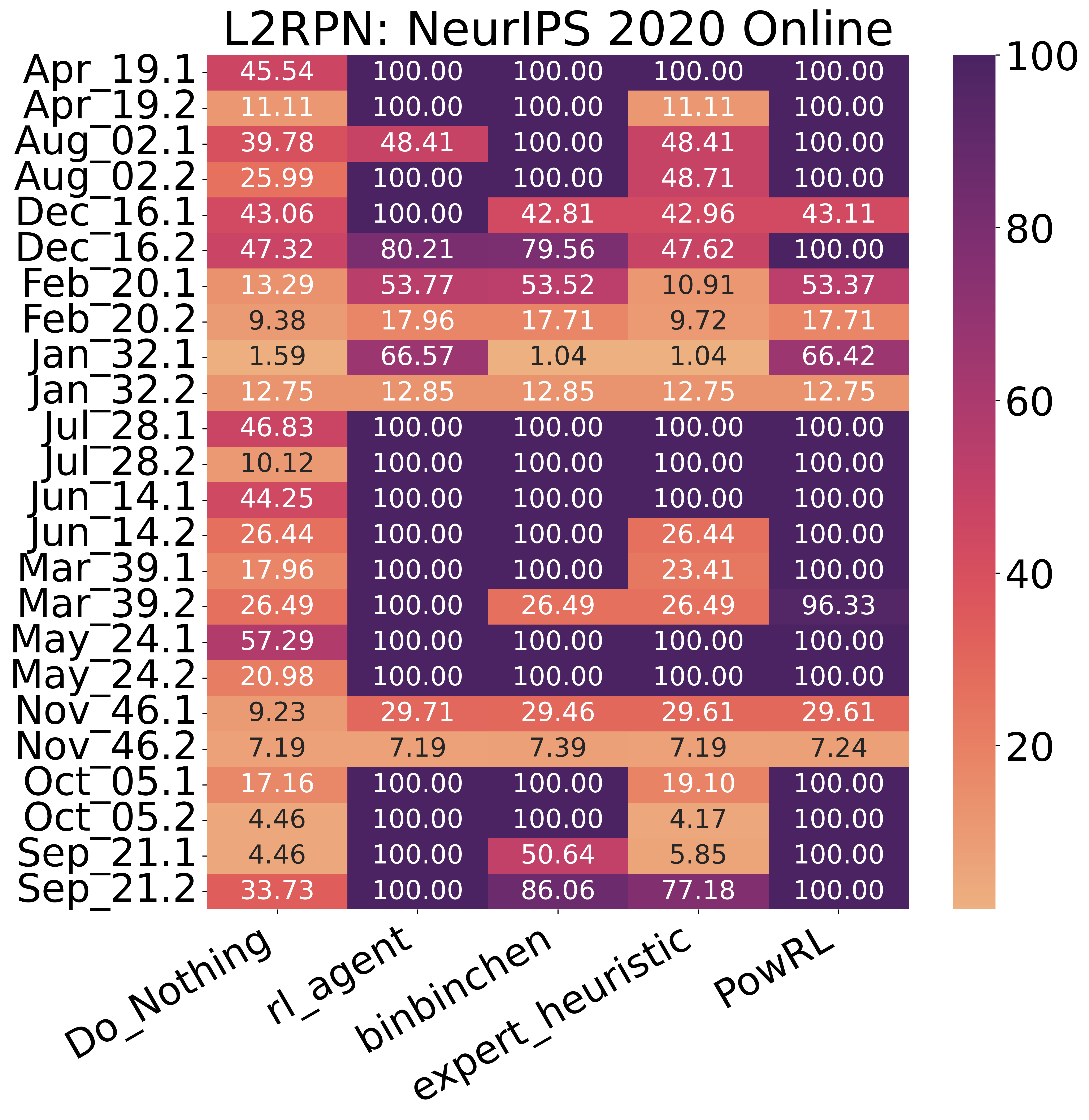} & \includegraphics[width=.82\columnwidth]{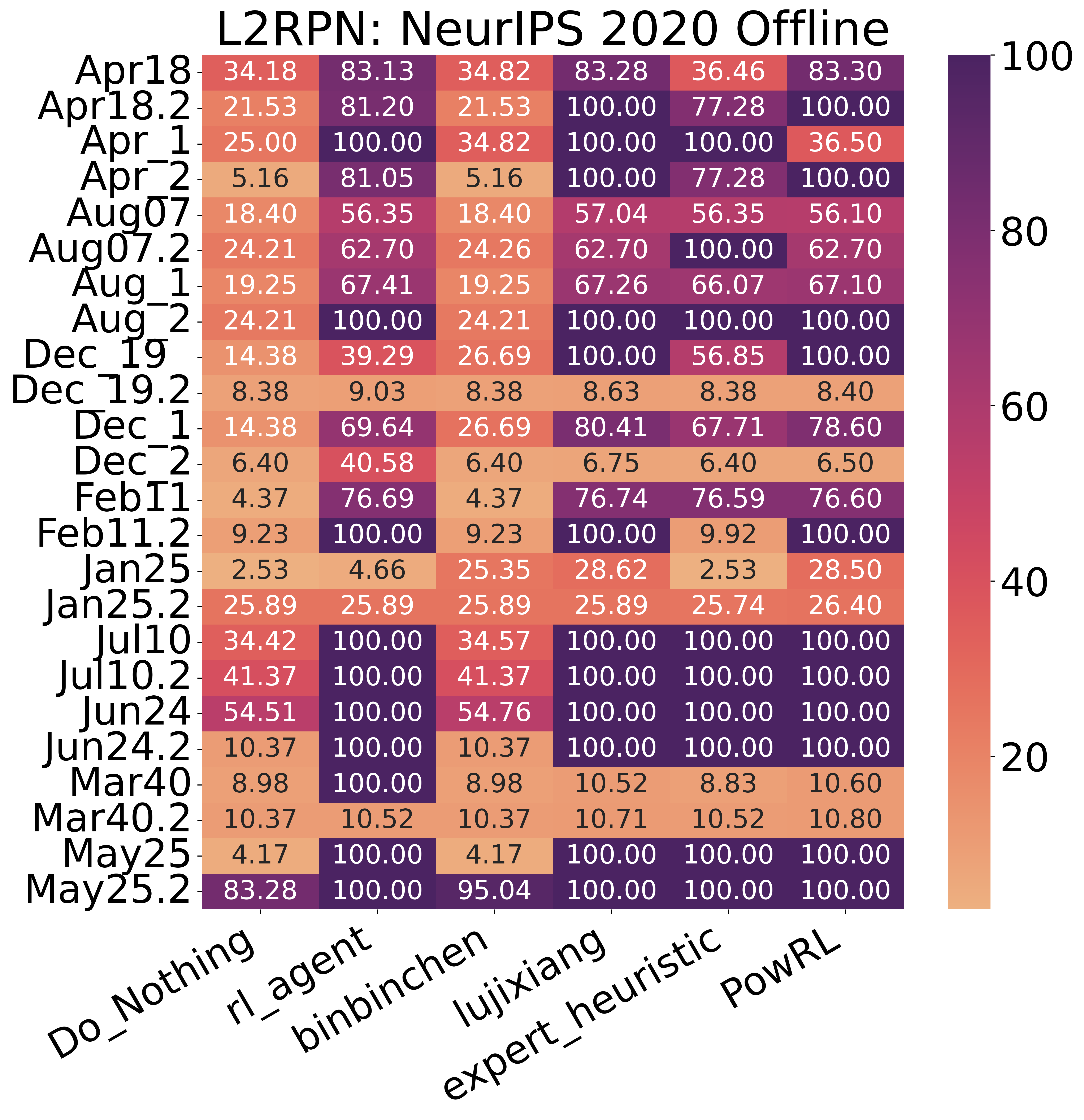}\cr
			(a) & (b)
		\end{tabular}
		\caption{Survival percentages of various agents on the L2RPN NeurIPS 2020 challenge dataset (a) Online, (b) Offline.}
		\label{fig:NeurIPS}
		\vspace{-.5em}
	\end{center}
	\vspace{-1.5em}
\end{figure*}

\noindent\textbf{Datasets} The dataset used for training PowRL is part of the small dataset included with the L2RPN NeurIPS 2020 challenge (Robustness track) starting kit~\cite{neurips}. It consists of power grid data worth 48 years spaced out at an interval of 5 minutes amounting to a total training data of 4,644,864 steps. The dataset is based on the reduced IEEE-118 system (see \figref{fig:PowGrid}) with 22 generators, 36 substations, 37 loads, and 59 power lines. some of the loads in this dataset represent interconnection with another grid. This dataset uses the Grid2Op framework to facilitate sequential decision-making process.

The L2RPN NeurIPS 2020 (Robustness track) uses two different datasets - (a) an \emph{offline} dataset included with the starting kit for participants to evaluate the performance of their trained models, (b) an \emph{online} dataset that is hidden from the participants. The participants can directly upload their trained models on the competition Codalab to get them evaluated and ranked. Both these datasets contain 24 episodic weekly scenarios resolved at a 5-minute interval. The hidden dataset was carefully picked to offer the different levels of toughness that might not have been observed during the training phase, even though both the datasets were generated from a similar statistical distribution. The previous version of this competition, the L2RPN WCCI 2020 challenge, differs only in terms of an adversary opponent. While the online submissions to the WCCI challenge are no longer available, an offline evaluation dataset is included with the WCCI starting kit~\cite{wcci}. This third test dataset consists of 10 different 3-day-long (864 steps) episodic scenarios, picked in order to offer different levels of difficulty. In this work, we benchmark the PowRL framework on all three publicly available datasets.

\noindent\textbf{Experimental setup} PowRL runs on Grid2Op, which uses the Gym interface to interact with an agent. This Grid2Op platform emulates the sequential decision-making in the power system, where each episode is divided into a list of states each corresponding to a time step of 5 minutes. Each of these states is described by the power flow at any given time step, which in turn is described by the amount of electricity supplied or consumed by the generators and loads, respectively. This data on power flow in each state is provided in the form of power network architecture, consumption, and generation by the RTE. We have also used the lightsim2grid~\cite{lightsim2grid} backend to the Grid2Op platform in order to accelerate the computation. The model was trained on a large shared facility comprising of 3 Tesla P100 GPUs, with a maximum job duration of 120 hours.

\noindent {\bf Neural network architecture and training} For the PowRL framework, an RL agent based on clipped PPO was trained on the L2RPN NeurIPS robustness track data. As the name suggests, the clipped PPO architecture~\cite{schulman2017proximal} clips the objective function (ratio between the new and the old policy scaled by the advantages), to remove incentives if the new policy gets too far from the older policy. The clipped PPO architecture consists of actor and critic networks; the actor produces the logits/ probability distribution of all possible actions, and the critic evaluates the goodness of the action suggested by the actor. For the actor network, we used four fully connected layers (with ReLU activation) of dimensions $[1000, 1000, 1000, 208]$. The output layer is a linear layer of a dimension equal to the number of discrete actions (2 out of 208 actions were deemed illegal since they bifurcated the network into two smaller groups). The critic network uses a single hidden layer (with ReLU activation) of dimension 64. The output dimension of the critic network is 1 and it uses the Gumbel-max trick to sample an action from the logits/probability distribution. The clipped PPO algorithm is implemented in PyTorch using Adam optimizer with a learning rate of  0.003, clip range of 0.2, rate/discount factor ($\gamma$) of 0.99, and generalized advantage estimation ($\lambda$) of 0.95. A single PPO agent was trained on the 36 parallel environments inside Grid2Op. The RL agent is engaged only when the maximum line overflow exceeds the safety threshold. Any such transition is stored in the prioritized replay buffer, and a long sequence of transitions is sampled (around 20k) in each epoch during the training.

\noindent\textbf{Baseline algorithms} We benchmark the PowRL framework against the other popular baseline approaches for power grid control. Readers are advised to refer to the section on Related Works for more details on the baseline algorithms.\\
\noindent{\bf 1. Do\_Nothing} The most basic of all algorithms, it contains only a single action of \texttt{do-nothing}, which is passed on to the environment at every step for all the scenarios.\\
\noindent{\bf 2. expert\_heuristic} This approach uses a combination of different actions collected based on the expert experience for the given power grid, where specific actions are taken in accordance with different situations faced.\\
\noindent{\bf 3. lujixiang}, the $3^{\text{rd}}$-place winner in NeurIPS 2020 challenge. This approach used `guided exploration', where the top-5 $Q$-value actions were sampled and simulated, and the action with best simulated reward was selected.\\
\noindent{\bf 4. binbinchen} the $2^{\text{nd}}$-place winner in NeurIPS 2020 challenge. It uses an imitation learning framework, where a simple feed-forward neural network (`Junior' student) is trained to imitate an expert agent, while a PPO-based `Senior' student is developed using the `Junior' student as the Actor.\\
\noindent{\bf 5. rl\_agent}, the winner of both the tracks in NeurIPS 2020 competition. It uses a black-box optimization approach to find optimal policies. In addition, for specific scenarios, such as in the month of March, the agent uses additional generator redispatch guided by the prior experience to survive some critical attacks in those scenarios.

% \begin{figure}
%     \centering
%     \includegraphics[width=.9\columnwidth]{Figures/wcci_offline_percentage.png}
%     \caption{Survival percentages of various agents on the L2RPN WCCI 2020 challenge (Offline).}
% 	\label{fig:WCCI}
% 	\vspace{-1.5em}
% \end{figure}
% \begin{figure}
%     \centering
%     \includegraphics[width=.96\columnwidth]{Figures/actions_per_substation.png}
%     \caption{Diversity of remedial actions distributed across multiple substations.}
% 	\label{fig:freq}
% 	\vspace{-1.5em}
% \end{figure}

\begin{figure*}
	\begin{center}
		\begin{tabular}{cc}
			\includegraphics[width=.8\columnwidth]{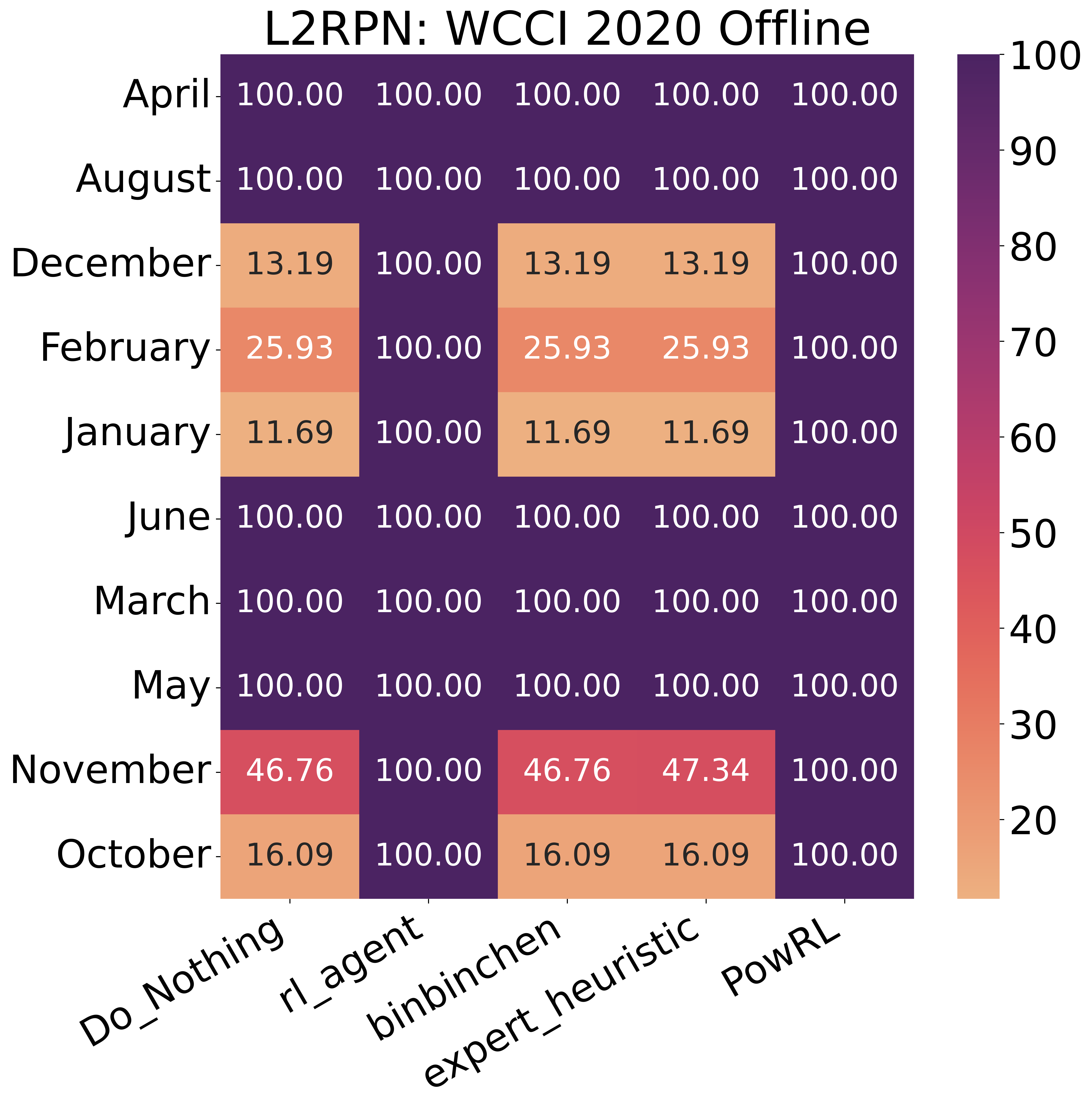} & \includegraphics[width=1.05\columnwidth]{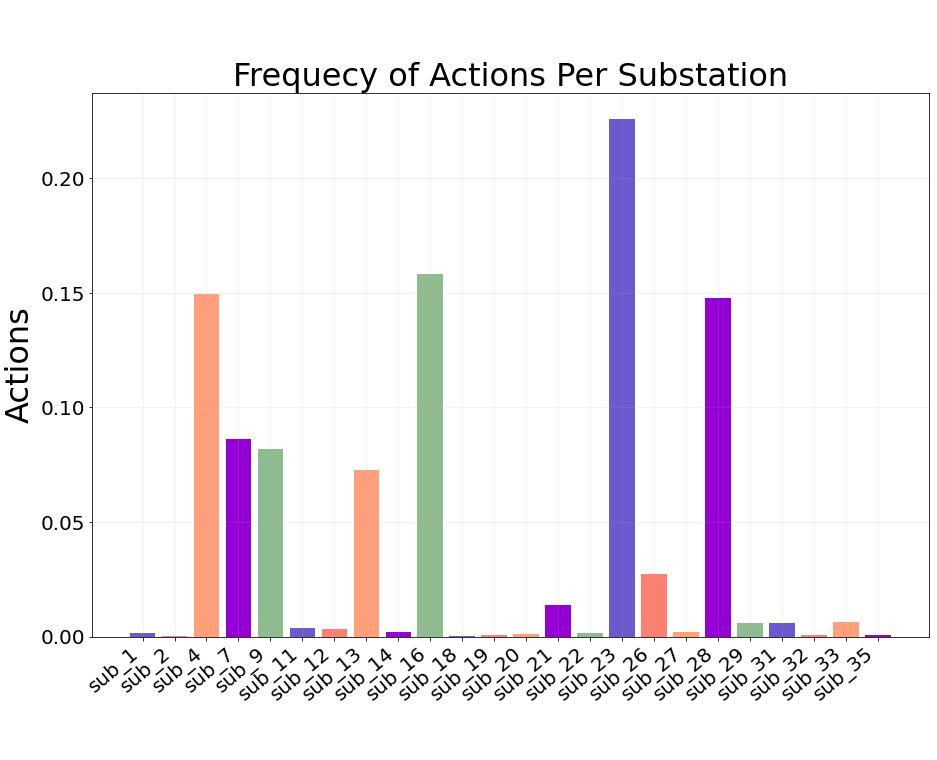}\cr
			(a) & (b)
		\end{tabular}
		\caption{(a) Survival percentages of various agents on the L2RPN WCCI 2020 challenge (Offline). (b) Diversity of remedial actions distributed across multiple substations.}
		\label{fig:WCCI}
		\vspace{-.5em}
	\end{center}
	\vspace{-1.5em}
\end{figure*}

\noindent\textbf{Results} The aim of the RL agent is to explore the flexibilities offered by cheap topology actions meant to support human decision-making operations. The performance of the proposed PowRL agent with baselines is evaluated over the three different test datasets mentioned above. Each scenario is evaluated/scored based on the operation cost and the losses due to blackout, hence we have used both survival steps and scenario score as the evaluation criterion.

We first benchmark the PowRL on NeurIPS 2020 online test dataset which is 24 different weekly episodic scenarios. \figref{fig:NeurIPS}a reflects the scenario-wise performance of our approach along with the other baselines. Both the PowRL and the rl\_agent fully survive the highest number of 16 scenarios, while the binbinchen and expert heuristics are able to survive fully only in 13 and 6 scenarios, respectively. It was found that a set of scenarios (January, February, November, December) looked tough, and required some new flexibilities, such as voltage control and load shedding, however, they were not made available in the environment. The dec\_16.2 scenario has one of the strongest power grid attacks~\cite{marot2021learning} and only our approach is able to survive successfully in this scenario. The other scenario is aug\_02.1 where the PowRL and binbinchen successfully survive the full episode. The other difficult scenario is mar\_39.2, where our agent survives nearly 96.3\% only through topological actions. Though the rl\_agent survives the full episode, it uses forced costly generator dispatch along with topology actions. Interestingly, while the rl\_agent, and the binbinchen were the eventual winners in the competition (announced in Nov'20), our approach now sits at the top of the challenge leaderboard, outperforming all other agents (see \tabref{tab:comp}).

We next evaluate the different approaches on NeurIPS 2020 offline test dataset (see \figref{fig:NeurIPS}b). Interestingly, the lujixiang agent, which was the eventual $3^{\text{rd}}$ place winner in the competition, survives the most 12 scenarios, while PowRL and the rl\_agent survive 11 and 10 scenarios, respectively. Note that the source code included with the lujixiang appears broken and hence, a scenario-wise analysis for the online data could not be obtained. However, the offline scenario-wise analysis was obtained from their talk during the winner announcement. Despite surviving a fewer number of scenarios, PowRL still outperforms the lujixiang agent indicating that our model requires less operational cost and tries to avoid losses due to blackouts. The PowRL is able to survive the highest number of steps in the dec\_2, feb11, and jan25.2 scenarios. In the Aug07.2 scenario, only the expert heuristic is able to survive the attack, as they are designed with the simple heuristic based on the generator dispatch and topological actions, rl\_agent deploys additional heuristic generator redispatch, which results in the best performance in the march scenario in comparison to other algorithms. The scenario named dec\_19 comprises of one of the strongest adversarial attacks and only PowRL survives it successfully. The binbinchen agent fares very poorly on this dataset and is unable to survive even a single scenario.

Finally, we evaluate the agents on the WCCI offline test dataset (see \figref{fig:WCCI}a). The PowRL agent and the rl\_agent successfully survive all the scenarios present in this dataset, which is expected since this test dataset aims primarily with congestion management (no adversaries). Nonetheless, it is interesting to note that the performance of the binbinchen agent drops significantly when tested on this test dataset, too, which clearly indicates overfitting issues during its training. All the results are summarized in \tabref{tab:comp}. It is also worth noting that the PowRL agent uses the least number of actions compared to other learning based agents. Despite using a much smaller action set, it is still able to achieve state-of-the-art-performance on all datasets. Additionally, the run-time with PowRL agent evaluated on the NeurIPS Online challenge is the least among all RL agents (see \tabref{tab:comp}). 

\noindent\textbf{Post-hoc analysis} We also analyzed the action sequences utilized by the PowRL agent on the NeurIPS online dataset. \figref{fig:WCCI}b shows the frequency with which an action at a substation is utilized by the PowRL agent in response to different attacks. We found out that the PowRL uses a total of 146 different actions spread across 25 different substations, which is notably diverse than the agents binbinchen (20 subs), rl\_agent (18 subs), and lujixiang (16 subs)~\cite{marot2021learning}. Evidently, the most frequent actionable substations are 4, 16, 23, and 28 because of the availability of large number of potential topological combinations. Additionally, the PowRL agent also engages substations 11, 18, 19, 20, and 22, though they have relatively fewer number of elements. Unlike the rl\_agent that encodes fixed sequences of actions on three substations that need to be acted on during instances of heavy overload, our PPO framework is capable of learning the optimal sequence combinations during the training phase itself. For instance, our analysis finds out the most frequently occurring combinations of substations during events of critical attacks in January: $[9\to13\to23\to16]$, $[7\to16\to23\to13]$, $[23\to9\to7\to13\to16]$, and in December: $[21\to23\to16]$.

\section{Conclusion} This study proposes a heuristic-guided RL framework, PowRL, for robust control of power networks subjected to production and demand uncertainty, as well as adversarial attacks. Using a careful action selection process, in combination with line reconnection and recovery heuristics, equips PowRL to outperform SOTA approaches on several challenge datasets even with reduced action space. PowRL not only diversifies its actions across substations, but also learns to identify important action sequences to protect the network against targeted adversarial attacks.

\bibliography{aaai23}
\end{document}